\documentclass{article}

\usepackage{arxiv}

\usepackage[utf8]{inputenc} 
\usepackage[T1]{fontenc}    
\usepackage{hyperref}       
\usepackage{url}            
\usepackage{booktabs}       
\usepackage{amsfonts}       
\usepackage{nicefrac}       
\usepackage{microtype}      
\usepackage{multirow}
\usepackage{array}
\usepackage{makecell}
\usepackage{graphicx}
\usepackage{comment}
\usepackage{latexsym}
\usepackage{graphicx}
\usepackage{subcaption}
\usepackage{graphics}
\usepackage{algorithm}
\usepackage{algpseudocode}
\usepackage{amssymb}
\usepackage{amsmath}
\usepackage{url}
\usepackage{tikz}
\usepackage{float}
\usepackage{enumitem}
\usepackage{color}

\title{SummPip: Unsupervised Multi-Document Summarization with Sentence Graph Compression\thanks{This work has been accepted to SIGIR 2020 for publication.}}

\author{
  Jinming Zhao \\
  Monash University\\
  Australia, VIC 3800 \\
  \texttt{jmzha2@student.monash.edu} \\
    \And
  Ming Liu \\
  Deakin University\\
  Australia, VIC 3217 \\
  \texttt{m.liu@deakin.edu.au} \\
    \And
  Longxiang Gao \\
  Deakin University\\
  Australia, VIC 3217 \\
  \texttt{longxiang.gao@deakin.edu.au} \\
  \And
 Yuan Jin \\
 Deakin University\\
  Australia, VIC 3217 \\
  \texttt{yuan.jin@deakin.edu.au}
  \And
  Lan Du \\
  Monash University\\
  Australia, VIC 3800 \\
  \texttt{lan.du@monash.edu} \\
    \And
  He Zhao \\
  Monash University\\
  Australia, VIC 3800 \\
  \texttt{ethan.zhao@monash.edu} \\
    \And
  He Zhang \\
  Shandesitong\\
  China, Beijing 100079 \\
  \texttt{zhanghe12@tsinghua.org.cn} \\
    \And
  Gholamreza Haffari \\
  Monash University\\
  Australia, VIC 3800 \\
  \texttt{Gholamreza.Haffari@monash.edu} \\
}

\begin{document}

\maketitle

\begin{abstract}
  Obtaining training data for multi-document summarization (MDS) is time consuming and resource-intensive, so recent neural models can only be trained for limited domains. In this paper, we propose SummPip:  an unsupervised method for multi-document summarization, in which we convert the original documents to a sentence graph, taking both linguistic and deep representation into account, then apply spectral clustering to obtain multiple clusters of sentences, and finally compress each cluster to generate the final summary. Experiments on Multi-News and DUC-2004 datasets show that our method is competitive to previous unsupervised methods and is even comparable to the neural supervised approaches. In addition, human evaluation shows our system produces consistent and complete summaries compared to human written ones.
\end{abstract}


\section{Introduction}
Text summarization (TS) aims to condense long documents into a few short sentences which cover the main themes of those documents. In general, there are two approaches to TS: extractive approaches, where salient pieces of text, e.g. words, phrases or sentences are identified and taken as the summary, and abstractive approaches, most of which rely on neural methods such as the Pointer-Generator Network \cite{see-etal-2017-get}. 
Different from single-document summarization, 
multi-document summarization (MDS) aims to effectively integrate key information from multiple text sources into a concise and comprehensive report.
%
State-of-the-art MDS systems are based on supervised learning, requiring relatively large amounts of labeled training data. 
However, obtaining  training data  is time consuming and resource-intensive. As a result, existing datasets are only available for limited domains. 

Recent years have witnessed an increasing number of summarization systems \cite{kryscinski-etal-2019-survey}. 
Variants of neural sequence-to-sequence models, which are originally developed for  machine translation  \cite{shi2018neural}, have been particularly successful in the summarization tasks. %
Despite the huge efforts of using deep neural models in summarization, they often require large-scale parallel corpora of input texts paired with their corresponding output summaries for direct supervision. 
For example, there are more than 280,000 training examples in the CNN/Daily Mail dataset\footnote{https://github.com/abisee/cnn-dailymail}, so it could be too costly to obtain ground-truth labels at this scale \cite{vu2019learning}. For most existing large datasets, and they consist of pairs of summary-like text and source text, for instance using news titles as summaries for news articles. 

In this case, unsupervised learning approaches are appealing as they do not require labeled data for summarization. They can be categorized into neural and non-neural approaches.
Neural summarization approaches utilize deep neural networks to generate summaries \cite{kryscinski-etal-2019-survey}. These models are often trained with autoencoders \cite{chu2018meansum} with a reconstruction loss. However, autoencoder approaches preserve every detail that helps to reconstruct the original documents, which is not applicable in the MDS settings. They also have limited capability in processing long text. Non-neural approaches tend to make the use of domain expertise in developing summarization systems \cite{kryscinski-etal-2019-survey}. In particular, current unsupervised approaches focus on building graphs to blend sentences from different documents and leveraging correlations among documents to extract the most representative sentences \cite{christensen-etal-2013-towards}. Yet, they are limited in incorporating finer linguistic information. 

Given that documents in MDS come from different sources, they are redundant and repetitive in expressing ideas, while summaries contain only a few key points. We believe semantic clusters can be formed according to the distances between sentences, 
and then each cluster can be compressed into a single sentence representing salient content. In this paper, we propose SummPip: an unsupervised multi-document summarization approach based on sentence graph compression.  To our best knowledge, we are the first to apply sentence compression to MDS. Our main contributions are:
\begin{itemize}
    \item SummPip is the first unsupervised summarization method which constructs sentence graphs by incorporating both linguistic knowledge and deep neural representations. It assumes that a summary sentence can be created by compressing a within-graph cluster. 
    \item We perform both automatic evaluation and human evaluation on two MDS benchmark datasets.
    SummPip shows significantly better performance than other unsupervised approaches and is compared favourably 
    to the latest supervised neural models.
\end{itemize}

\section{SummPip: Automatic pipeline for Multi-Document Summarization}

\begin{figure}[ht]
  \centering
  \includegraphics[width=0.8\linewidth]{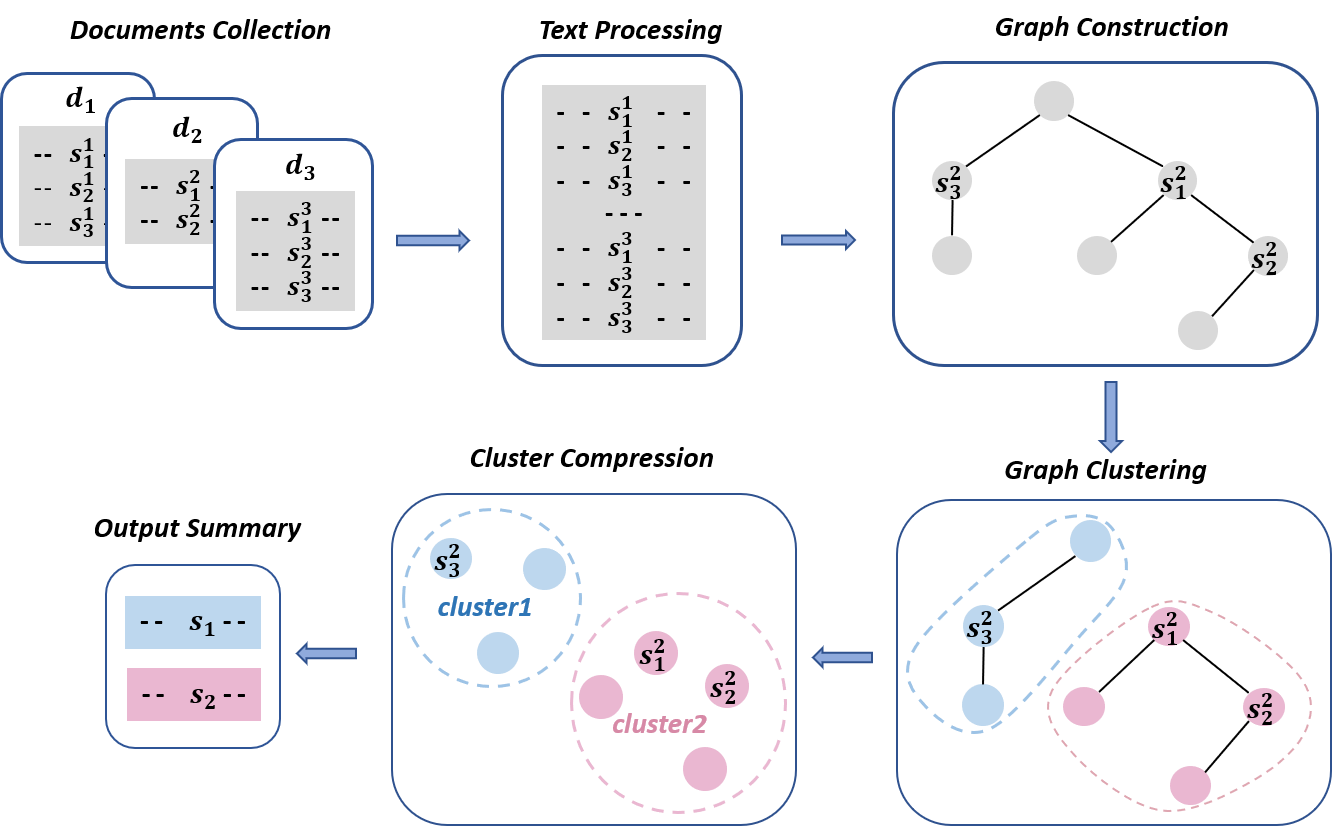}
  \vspace{4mm}
  \caption{Automatic pipeline for unsupervised multi-document summarization.}
  \label{fig:arc}
\end{figure}

Our pipeline consists of four major steps:  i) Conduct document processing. ii) Build a structured sentence graph, where the nodes correspond to the sentences generated at step one and the edges are drawn based on both the lexical and the deep semantic relations between sentences. iii) Apply graph clustering to get within-graph partitions. iv) Generate summary texts from the extracted sub-graphs. Figure \ref{fig:arc} illustrates our pipeline for unsupervised MDS. We specify each step in the following subsections\footnote{
Our code is available at https://github.com/mingzi151/SummPip}.
\subsection{Text Processing} 
Given a set of theme-related documents $D= {d_1,d_2,...,d_n}$, our goal is to generate a summary $S$ which is concise and faithful to the original documents. We concatenate documents $D$ and apply minimal text processing, mainly sentence split, as we want to keep documents raw for subsequent processing. We keep all the stop words for the last compression step. The processing steps are conducted with SpaCy NLP modules. The output here is a list of sentences, which are fed to the sentence graph construction step.

\subsection{Sentence Graph Construction}
The goal at this step is to identify pairwise sentence connection which represents the discourse structure of documents $D$. We construct a sentence graph based on Approximate Discourse Graph (ADG) \cite{christensen-etal-2013-towards} together with deep embedding based techniques. Specifically, we build a  graph $(V,E)$, where each node $v_i \in V$ represents a sentence, and nodes $v_i$ and $v_j$ ($i\neq j$) are connected, i.e., their edge $e_{i,j} = 1$, if one of the following conditions is met. i) \textbf{Deverbal Noun Reference}: It is common that when an event is mentioned in a verb phrase, it is referred by a deverbal noun or noun phrase in the subsequent sentence. To recognize such a relationship, we derive the noun form of the verb phrase with \textit{derivationally\_related\_forms} links of WordNet before obtaining its semantically similar words by leveraging learned representations of words in a multi-dimensional space.
ii) \textbf{Entity Continuation}: The second condition concerns lexical chains for the purpose of local coherence. We insert an edge from sentence $v_i$ to sentence $v_j$ if they contain the same entity type (e.g., organization, person, product).
iii) \textbf{Discourse Markers}: We use a set of discourse markers (e.g., however, meanwhile, furthermore) to discover discourse relations between two adjacent sentences of a document. iv) \textbf{Sentence Similarity}: We obtain sentence representations by averaging all of the word vectors of a sentence. Sentence similarity score is calculated by taking cosine similarity of two sentence vectors.

\subsection{Spectral Clustering on Sentence Graph}
Most graph clustering approaches try to identify communities of nodes in a graph based on the edges linking them. We use spectral clustering as follows: we first get the Laplacian matrix based on the above sentence graph, and compute the first $k$ eigenvectors of the Laplacian matrix to define a feature vector for each sentence. Then we run k-means on these features to separate those sentences into $k$ classes.

\subsection{Multi-sentence Compression for Summary}
Multi-sentence compression (MSC) generates a single summary sentence from each cluster that contains a set of semantic related sentences. We use MSC as the last step of our method to generate summaries from the sentence clusters.
A typical approach \cite{filippova2010multi} is to build a word graph and take the shortest path of the words as the summary. 
We use an extended version of this method \cite{boudin2013keyphrase} by considering keyphrases to adjust the compression process, so that word paths with keyphrases are given higher scores than other word paths. It yields a few compressed summaries from the sentence clusters, in which we select the summary with the highest score as our final output. 

\begin{table*}
\begin{center}
  \caption{Experiment Results on Multi-News and DUC2004. \textbf{Bold}: Better in the unsupervised setting.  \underline{Underlined}: Best result among both supervised and unsupervised settings. $^*$: Statistically significantly better than Centroid with p-value $\le .05$.}
  \scalebox{0.9}{
  {%
  \begin{tabular}{lcccccccc}
    \toprule
    \multirow{2}{*} & 
    &   Multi-News & &  &DUC-2004 &\\
    Model & R-1 & R-2 & R-SU & R-1 & R-2 & R-SU \\
    \midrule
    \textbf{Lead-3} & 39.41 & 11.77 & 14.51 & 30.77 & 8.27 & 7.35  \\
    \hline
    \textit{Unsupervised}\\
    \textbf{LexRank} \cite{erkan2004lexrank} (reproduced) & 38.27 & 12.70 & 13.20 & 35.56 & 7.87  & 11.86\\
    \textbf{TextRank} \cite{mihalcea2004textrank} & 38.44 & 13.10 & 13.50 & 33.16 & 6.13 & 10.16  \\
    \textbf{MMR} \cite{carbonell1998use} & 38.77 & 11.98 & 12.91 & 30.14 & 4.55 & 8.16 \\
    \textbf{Centroid} \cite{rossiello2017centroid} (reproduced)& 41.25 & 12.83 & 15.63 & 36.52 & 8.82 & 11.68  \\
    \hline
    \textit{Supervised}\\
    \textbf{PG-Original} \cite{see-etal-2017-get} & 41.85 & 12.91 & 16.46 & 31.43 & 6.03 & 10.01  \\
    \textbf{PG-MMR} \cite{lebanoff-etal-2018-adapting} & 40.55 & 12.36 & 15.87 & \underline{36.42} & \underline{9.36} & \underline{13.23}  \\
    \textbf{Copy-Transformer} \cite{gehrmann-etal-2018-bottom} & \underline{43.57} & 14.03 & 17.37 & 28.54 & 6.38 & 7.22  \\
    \textbf{Hi-MAP} \cite{fabbri-etal-2019-multi} & 43.47 & \underline{14.89} & \underline{17.41} & 35.78 & 8.90 & 11.43  \\
    \hline
    \textbf{Our method} & \textbf{42.32}$^*$ & \textbf{13.28}$^*$ & \textbf{16.20}$^*$ & 36.30 & 8.47 & 11.55  \\
    \bottomrule
  \end{tabular}}
  }
  \label{tab:results}
\end{center}
\end{table*}

\section{Experiments}
In our experiments, we use ROUGE metrics \cite{lin2004rouge} to quantify the performance of SummPip on Multi-News and DUC-2004.

\textbf{Data sets} We experiment on Multi-News \cite{fabbri-etal-2019-multi}, which is a large-scale dataset for MDS. The training data is only used to learn word vectors, and evaluation is done on test data which has 5,622 collections of documents. We truncate the multi-news documents size to 500 tokens, since the experiments on MDS suggest that increasing document length does not show significant performance improvement \cite{fabbri-etal-2019-multi}. We also apply our method to DUC-2004 \cite{over2004introduction}, which is a benchmark MDS dataset. Each instance is truncated to 1500 tokens. Detailed statistics of the datasets are reported in Table \ref{tab:data}.
\begin{table}[ht]
\begin{center}
    \caption{Dataset statistics}
    \scalebox{0.9}{
    \begin{tabular}{ c | c | c | c | c }
    \hline
    Datasets & test size & \thead{\# words \\ (doc)}  &  sources & 
    \thead{\# words \\(summary)} \\
    \hline
    Multi-News & 5622 & 2103.49  & 2.76 & 263.66  \\
    DUC-2004 & 50 & 5978.2   & 10 & 107.04  \\
    \hline
    \end{tabular}}
    \label{tab:data}
\end{center}
\end{table}

\textbf{Experiment Settings} We set the number of most similar words for deverbal nouns as 10. To detect discourse relations between adjacent sentences, we arbitrarily define 39 discourse markers and use them as signals. The threshold for sentence similarity is 0.98. We employ the SpectralClustering method from sklearn\footnote{https://scikit-learn.org/stable/modules/generated/sklearn.cluster.SpectralClustering.html} for graph clustering. The number of clusters is the average number of sentences in the reference summaries, which is 9 for Multi-News and 7 for DUC-2004. At the cluster compression step, we set the hyper-parameter $\alpha$ for the minimal number of words for the best compression as 5 for Multi-News and 14 for DUC-2004. We also conduct experiments with weighted sentence graphs, where the weights are the similarity scores between two sentences. However, our experimental results show no improvement compared to the unweighted graphs.

\textbf{Baselines} We compared our method with the following baselines:
Lead-3 is the baseline method for many news summarization tasks since reaserchers have found that the news summary is biased towards the beginning part. LexRank \cite{erkan2004lexrank}, TextRank \cite{mihalcea2004textrank}, MMR \cite{carbonell1998use} and Centroid \cite{rossiello2017centroid} are unsupervised methods, while the first two are graph based and the other two are similarity based approaches. We also bring in several strong supervised baselines, including PG-Original \cite{see-etal-2017-get}, PG-MMR \cite{lebanoff-etal-2018-adapting}, Copy-Transformer \cite{gehrmann-etal-2018-bottom} and Hi-MAP \cite{fabbri-etal-2019-multi}. All of them are neural summarization models, among which Hi-MAP is the latest reported neural method on Multi-News.

\textbf{Results} Table \ref{tab:results} shows the ROUGE results on Multi-News and DUC-2004. Our method produces significantly better results than previous unsupervised methods on Multi-News, including the strong Centroid baseline, and being highly competitive to supervised deep neural models. For DUC-2004, SummPip also exhibits good performance, even though it does not outperform Centroid, which extract the whole sentences. Meanwhile, our method shows very close performance compared to the supervised neural models on both datasets, without any heavy learning process.

\section{Human Analysis}

We conduct human evaluation to further evaluate our method. We consider the following aspects of the generated summary. \textbf{Fluency}: there are two aspects regarding fluency, i.e., within-sentence fluency and coherence of sentences.
\textbf{Consistency}: this reflects whether the system summary is consistent with the source text.
\textbf{Coverage}: this aspect indicates whether the shortened text covers  
the most salient information in the source documents.
\textbf{Redundancy}: it qualifies how well a summary is written or generated in reducing redundant information. A high-quality summary should not contain detailed description or repetitive information.
We instruct the human annotators to read 75  randomly sampled documents and rate reference summaries and system summaries on an 1-5 point scale (1 is the worst and 5 is the best). The annotators have no knowledge about which ones are written by human. They are also asked to give their preference for the paired summaries. 

\begin{figure*}
  \centering
  \includegraphics[width=0.9\linewidth,scale=0.1]{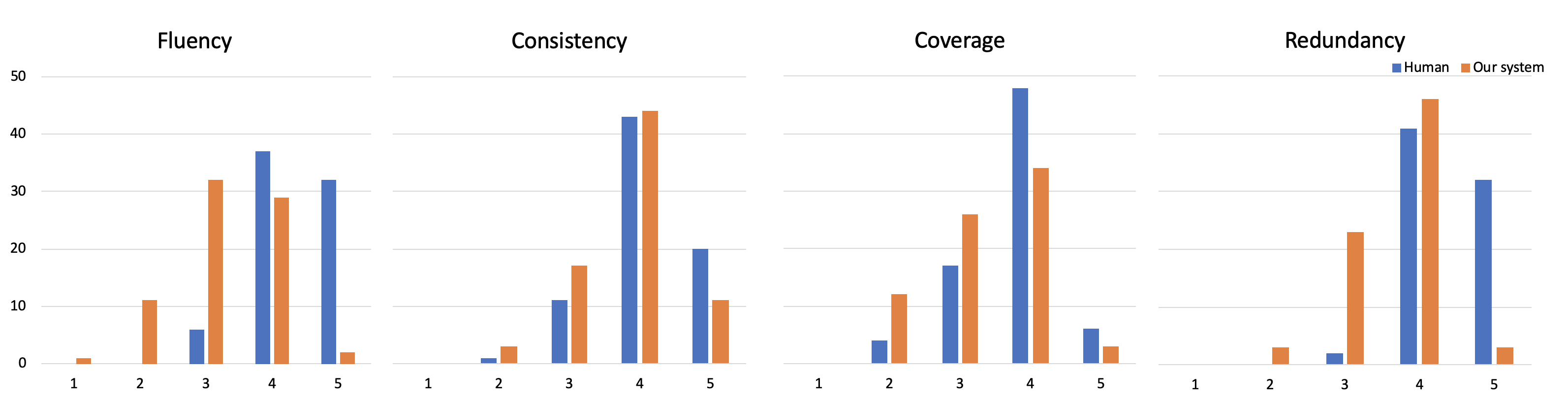}
  \caption{Human evaluation results on human-written summaries and SummPip summaries. The horizontal axis is the score, the vertical axis is the number of votes.}
  \label{fig:flow}
\end{figure*}

\begin{figure}[ht]
  \centering
  \includegraphics[width=0.75\linewidth]{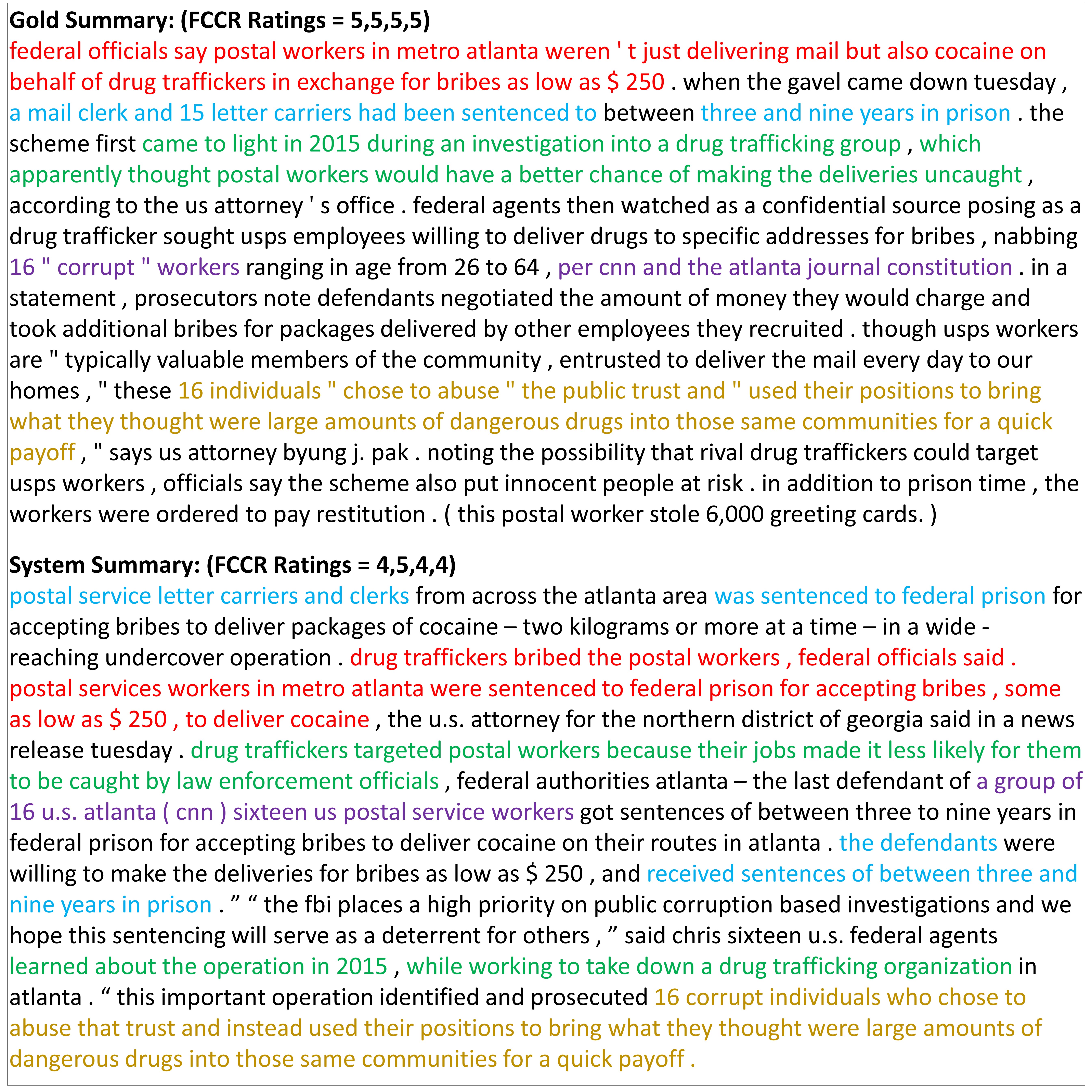}
  \caption{Example of a gold summary and a system summary.}
  \label{fig:example}
\end{figure}

Figure 2 shows the results regarding the above criteria. Our summaries are found to highly coincide with manual summaries in consistency and coverage, while showing less coincidence with manual summaries in fluency and redundancy. This implies that SummPip is capable of detecting important content in documents and generating accurate summaries, similar with the contents written by human. The annotators commented that in most cases, it was obvious to them which summary was generated by the machine. This is also reflected in the low number of score 5 in fluency and redundancy. Having said that, the peaks in the upper scoring area for our model, i.e., score 3 and 4, reveal that our model does not compromise much readability and conciseness of summaries. After inspecting the intermediate outcomes of our summaries, we find the redundancy in our summaries is caused by the occasional failure of the model in splitting sentences, which hurts the subsequent stages in the pipeline. We believe fluency and redundancy scores will increase if we improve the text processing step; we will leave this for further study. In addition to this, there were 13 votes for our summaries and 62 votes for gold summaries. This isn't surprising to us; rather, it indicates SummPip's great potential in producing high-quality summaries. Figure \ref{fig:example} is an example of a generated summary by SummPip, compared to the reference summary\footnote{The source documents can be found at: https://variety.com/2017/tv/news/jon-stewart-stephen-colbert-donald-trump-late-show-1201975014/; https://www.nydailynews.com/entertainment/tv/jon-stewart-blasts-trump-presidency-vindictive-chaos-article-1.2961096}. FCCR is short for the above four criteria. Each piece of salient information in the gold summary is given different colors to highlight their presence in the system summary.

\section{Conclusion}
We have proposed SummPip, a novel unsupervised method for multi-document summarization, in which we build a sentence graph, apply graph clustering and sentence compression to obtain partitioned summaries. Notably, the main novelties of SummPip are the combinations of sentence graphs and sentence compression, which, to the best of our knowledge, are leveraged for the first time in the literature of unsupervised multi-document summarisation.
Experiments on Multi-News and DUC-2004 show our method is effective and even comparable to the strong supervised neural approaches. Moreover, human evaluation indicates the great potential of our approach in producing high-quality summaries. In the future, we plan to add a ranking mechanism to the sentence graph construction stage and apply autoencoders to get more abstractive summaries during the compression step.

\bibliographystyle{unsrt}  

\bibliography{references}

\end{document}